\documentclass[11pt,a4paper]{article}
\usepackage[hyperref]{acl2019}
\usepackage{times}
\usepackage{latexsym}
\usepackage{booktabs}

\usepackage{url}

\aclfinalcopy 

\title{Partially Shuffling the Training Data to Improve Language Models}

\author{
	Ofir Press \\
	Paul G. Allen School of Computer Science \& Engineering,
	University of Washington \\
	{\tt ofirp@cs.washington.edu }
}
\date{}

\begin{document}
\maketitle
\begin{abstract}
Although SGD requires shuffling the training data between epochs, currently none of the word-level language modeling systems do this. Naively shuffling all sentences in the training data would not permit the model to learn inter-sentence dependencies. Here we present a method that partially shuffles the training data between epochs. This method makes each batch random, while keeping most sentence ordering intact. It achieves new state of the art results on word-level language modeling on both the Penn Treebank and WikiText-2 datasets.\footnote{Our code is available at {\tiny \url{https://github.com/ofirpress/PartialShuffle}}}
\end{abstract}

\section{Background}
A language model is trained to predict word $n+1$ given all previous $n$ words. A recurrent language model receives at timestep $n$ the $n$th word and the previous hidden state and outputs a prediction of the next word and the next hidden state. 

The training data for word-level language modeling consists of a series of concatenated documents. The sentences from these documents are unshuffled. This lets the model learn long term, multi-sentence dependencies between words. 

The concatenation operation results in a single long sequence of words. 
The naive way to train a language model would be to, at every epoch, use the entire training sequence as the input, and use the same sequence shifted one word to the left as target output.
Since the training sequence is too long, this solution is infeasible.

To solve this, we set a back propagation through-time length ($b$), and split the training sequence into sub-sequences of length $b$. In this case, in each epoch the model is first trained on the first sub-sequence, and then on the second one, and so on. While gradients are not passed between different sub-sequences, the last hidden state from sub-sequence $m$ becomes the initial hidden state while training the model with sub-sequence $m+1$. 

For example, if the training sequence of words is:

\texttt{[A B C D E F G H I J K L]}

\noindent for $b = 3$, the resulting four sub-sequences are:

\texttt{[A B C] [D E F] [G H I] [J K L]}

Note that we only present the input sub-sequences, as the target output sub-sequences are simply the input sub-sequences shifted one word to the left\footnote{For example, the target output sub-sequences here are \texttt{[B C D] [E F G] [H I J] [K L *]}, where \texttt{*} is the end-of-sequence token.}. 
This method works, but it does not utilize current GPUs to their full potential. 

In order to speed up training, we batch our training data. We set a batch size $s$, and at every training step we train the model on $s$ sub-sequences in parallel. 

To do this, we first split the training sequence into $s$ parts. Continuing the example from above, for $s=2$, this results in:

\texttt{[A B C D E F]}

\texttt{[G H I J K L]}

Then, as before, we split each part into sub-sequences of length $b$:

\texttt{[A B C] [D E F]}

\texttt{[G H I] [J K L]}

\noindent Then, during the first training step in each epoch we train on:

\texttt{[A B C] }

\texttt{[G H I] }

\noindent and during the second training step in each epoch we train on:

\texttt{[D E F] }

\texttt{[J K L]}

Note that at every step, all sub-sequences in the batch are processed in parallel. 

Before we introduced batching, in each epoch the output for each word in the training sequence was dependant on all previous words. 
With batching, the output of the model for each word is only dependant on the previous words in that batch element (or equivalently, row in our example), and the other words are ignored.

In our example, the hidden state that is given when inputting \texttt{G} is the default initial hidden state, and not the one that resulted after the input of \texttt{F}. This is not optimal, but since batching reduces the training time by a significant amount, all current models use this method. 

\section{The Partial Shuffle Method}
While SGD calls for random batches in each epoch, in existing language models, the data is not shuffled between epochs during training. This means that batch $i$ in every epoch is made up of the same sub-sequences.

The straightforward way to shuffle the data would be to shuffle all sentences in the training sequence between each epoch. This hurts the language model's performance, since it does not learn inter-sentence dependencies.  

Here we present the Partial Shuffle method, which improves the performance of the model.

Like before, we first separate the sequence of words into $s$ rows. Using the example sequence from above, this would result in (for $s=2$):

\texttt{[A B C D E F]}

\texttt{[G H I J K L]}

Then, for each row, we pick a random index between zero and the length of the row and we take the words that are located before this index and move them to the end of the row. So in our example, if the random index for row one was $2$ and for row two was $5$ this would result in (red marks the words which were moved):

\texttt{[C D E F {\color{red} A B}]}

\texttt{[L {\color{red}G H} {\color{red}I J K}]}

Finally, as before, each row (or equivalently, batch element) is divided into back-propagation through time segments. For $b = 3$, this will result in:

\texttt{[C D E] [F A B]}

\texttt{[L G H] [I J K]}

This method randomizes the batches while still keeping most of the word ordering intact. 
\begin{table}[t!]
\begin{center}
{\small
\begin{tabular}{@{}lcc@{}}

\toprule
Model                                & Validation & Test  \\ \midrule
MoS                                  & 58.27      & 56.18 \\
MoS + Partial Shuffle              & 57.43      & 55.35 \\
MoS + Finetune                       & 56.76      & 54.64 \\
MoS + Finetune + Partial Shuffle   & 55.89      & 53.92 \\ \midrule
DOC                                  & 55.39      & 53.44 \\
DOC + Partial Shuffle              & 54.90      & 53.28 \\
DOC + Finetune                       & 54.62      & 52.87 \\
DOC + Finetune + Partial Shuffle   & 54.30      & 52.58 \\
DOC + Finetune$^*$                    & 54.18      & 52.38 \\
DOC + Finetune$^*$ + Partial Shuffle & 53.79      & 52.00 \\ \bottomrule

\end{tabular} }
\end{center}
\caption{\label{PTB} Model perplexity on the Penn Treebank, without and with the Partial Shuffle method. Finetune$^*$ denotes repeating the finetuning operation three times.  }
\end{table}

\begin{table}[t!]
\begin{center}
{\small

\begin{tabular}{@{}lcc@{}}
\toprule
Model                                & Validation & Test  \\ \midrule
MoS                                  & 65.94      & 63.35 \\
MoS + Partial Shuffle              & 64.09      & 61.97 \\
MoS + Finetune                       & 63.98      & 61.49 \\
MoS + Finetune + Partial Shuffle   & 62.38      & 59.98 \\ \midrule
DOC                                  & 61.68      & 59.64 \\
DOC + Partial Shuffle              & 61.28      & 58.93 \\
DOC + Finetune                       & 60.97      & 58.55 \\
DOC + Finetune + Partial Shuffle   & 60.58      & 58.20  \\
DOC + Finetune$^*$                     &    60.29        &   58.03    \\
DOC + Finetune$^*$ + Partial Shuffle &  60.16         &  57.85     \\ \bottomrule
\end{tabular} }

\end{center}
\caption{\label{PTB} Model perplexity on WikiText-2, without and with the Partial Shuffle method. Finetune$^*$ denotes repeating the finetuning operation three times. }
\end{table}

\section{Results}
We evaluate our method on the current state of the art model, DOC~\cite{doc}, and the previous state of the art model, MoS~\cite{mos}, on the Penn Treebank~\cite{ptb} and WikiText-2~\cite{wt2} language modeling datasets. For each model, the hyper-parameters (including $b$ and $s$) are not modified from their original values. In addition, we present results for finetuned~\cite{awd} models, with and without the Partial Shuffle. 

Our shuffling method improves the performance of all models, and achieves new state of the art results on both datasets. Our method does not require any additional parameters or hyper-parameters, and runs in less than $\frac{1}{100}$th of a second per epoch on the Penn Treebank dataset.

\section{Acknowledgements}
This note benefited from feedback from Judit Acs, Shimi Salant and Noah A. Smith, which is acknowledged with gratitude.

\bibliography{acl2019}
\bibliographystyle{acl_natbib}

\end{document}